\pdfoutput=1

\documentclass[11pt]{article}

\usepackage{acl}

\usepackage{times}
\usepackage{latexsym}

\usepackage[T1]{fontenc}

\usepackage[utf8]{inputenc}

\usepackage{microtype}

\usepackage{graphicx} 
\usepackage{inconsolata}
\usepackage{algorithm}
\usepackage{algpseudocode}
\usepackage{amsmath}
\usepackage{multirow}
\usepackage{xcolor}
\usepackage{colortbl}
\usepackage{array}
\usepackage{booktabs}
\usepackage{subcaption}
\usepackage{newfloat}
\usepackage{listings}
\usepackage{rotating}
\usepackage{amsfonts} 

\definecolor{purple}{HTML}{6600CC}  
\title{Towards Faithful Knowledge Graph Explanation Through Deep Alignment in Commonsense Question Answering}

\author{Weihe Zhai \\
  Harbin Institute of Techology\\
  \texttt{weihezhai@insun.hit.edu.cn} \\\And
  Arkaitz Zubiaga \\
  Queen Mary Univerisity of London\\
  \texttt{a.zubiaga@qmul.ac.uk} \\  }

\begin{document}
\maketitle
\begin{abstract}
The fusion of language models (LMs) and knowledge graphs (KGs) is widely used in commonsense question answering, but generating faithful explanations remains challenging. Current methods often overlook path decoding faithfulness, leading to divergence between graph encoder outputs and model predictions. We identify confounding effects and LM-KG misalignment as key factors causing spurious explanations. To address this, we introduce the LM-KG Fidelity metric to assess KG representation reliability and propose the LM-KG Distribution-aware Alignment (\textit{LKDA}) algorithm to improve explanation faithfulness. Without ground truth, we evaluate KG explanations using the proposed Fidelity-Sparsity Trade-off Curve. Experiments on CommonsenseQA and OpenBookQA show that LKDA significantly enhances explanation fidelity and model performance, highlighting the need to address distributional misalignment for reliable commonsense reasoning.
\end{abstract}

\section{Introduction}
\begin{figure}[t]
    \includegraphics[width=\columnwidth]{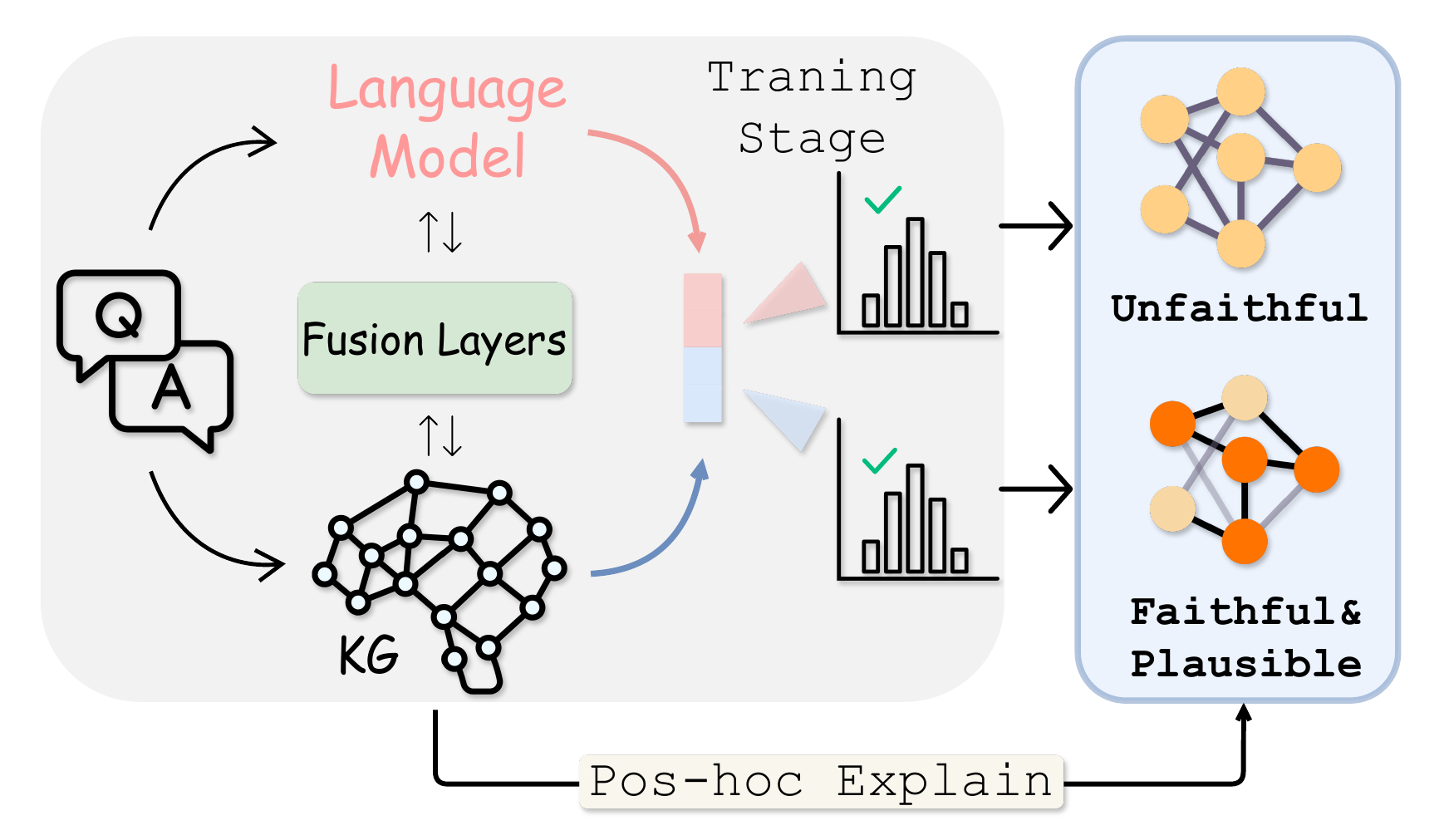}
\caption{This figure depicts a class of models that integrate KG and LM for question answering. The training stage on the left side of the figure mainly includes LM, KG, and their interaction through a knowledge exchange fusion layer. The right side of the figure illustrates the post-hoc explanation results. Explanations extracted from the KG of models that produce the same correct answers can be inconsistent and unfaithful.}
    \label{exp}
\end{figure}

In commonsense reasoning problems, many rely on both explicit textual information and structured domain knowledge \citep{hirschman2001natural} to compensate for the limited factual memory of LMs \cite{li2022pre} and provide insights into the inference processes \cite{danilevsky2020survey}, however explanations can also be expressed by highlighting a subset of this knowledge. Making the model output the facts used to answer a particular question can increase trustworthiness and help with debugging. Effective explanations should accurately reflect the reasoning process of a model \cite{herman2017promise}. In knowledge-augmented commonsense QA, attention weights from message-passing have been used to provide poc-hoc explanations \cite{lin2019kagnet, yasunaga2021qa}, as illustrated in Figure \ref{exp}. However, the reliability of these explanations has been questioned \cite{jain-wallace-2019-attention}, and the criteria for evaluating model explainability are often neglected, diminishing their impact.

We argue that explanations from a broad class of KG-enhanced LMs (LM-KG) are of limited faithfulness. The behaviour of graph encoder deviates from the overall LM-KG model and it has limited influence on the prediction, so explanations extracted from the graph encoder are unlikely to reflect the full set of facts. Besides, this process does not guarantee that the extracted explanations will be faithful to the reasoning of the model \cite{wiegreffe-pinter-2019-attention}, leading to what we call spurious explanations\cite{zhao2023towards}.

Spurious explanations, which lie outside the genuine rationale of the model's prediction, can arise due to various factors. The Graph Neural Network (GNN) learned from the knowledge graph may preserve the prediction but deviate from the original model's reasoning due to confounding effects. In LM-KG models, the LM compensates for the reasoning of the weakly-trained GNN, making it more vulnerable to such issues. Consequently, the extraction of explanations becomes unreliable.


To address these challenges, we make the following contributions:
\begin{enumerate}
\item We measure model faithfulness by deeply detaching the LM's ability to influence the final prediction, providing a design method for faithful models that can generalise to similar architectures.
\item We analyse the underlying mechanism of spurious explanations and discuss why graph motifs (structure) can enhance model performance but fail to produce faithful explanations.
\item We propose the \textbf{LM-KG Fidelity} and \textbf{LM-KG Consistency} metrics, which directly inspire the development of the \textbf{LM-KG Distribution-aware Alignment} (\textit{LKDA}) training architecture.
\item We introduce a joint Fedility-Sparsity measurement method to help analyse whether the attention weights of the GNN contain explanatory paths.
\end{enumerate}
Our analysis, conducted on the CommonsenseQA and OpenBookQA datasets, demonstrates that \textit{LKDA} enhances KG fidelity across various LM-KG models, representing a significant contribution to graph explainability and setting a new benchmark for future research. Furthermore, \textit{LKDA} consistently improves the overall performance accuracy of models. On the OpenBookQA dataset, some models exhibit an accuracy increase of approximately 10\% while maintaining the same model architecture and parameter count. These suggest that our proposed method can assist models in better utilising the structured knowledge contained within the Knowledge Graph.

\section{Related Work}
\subsection{Knowledge Graphs in NLP} Research has explored enhancing NLP with additional knowledge. Studies have shown pre-trained language models can serve as implicit knowledge bases \cite{pan2019improving, petroni2019language}. Others have integrated structured knowledge graphs into language models for better knowledge representation, focusing on processing the knowledge graph (KG) and the language model (LM) separately before combining them for question answering (QA) tasks \cite{mihaylov2018knowledgeable, wang2019improving, zhang2022greaselm, lin2019kagnet, yasunaga2021qa}.

\subsection{Multi-relational Graph Encoder} Graph Neural Networks (GNNs) are significant in handling diverse graph structures \cite{kipf2017semisupervised,veličković2018graph}. For multi-relational graphs like KGs, which have complex relational data, R-GCNs and GAT have been developed to handle these relations effectively \cite{schlichtkrull2018modeling, veličković2018graph}.

\subsection{KGs for Post-hoc Explanations in LMs} 
LMs struggle with interpretability \cite{danilevsky2020survey}. Grounding LM outputs in KGs has been a method to provide explanations, but these are often not fully representative due to the reliance on text and graph embeddings \cite{feng2020scalable, sun2022jointlk, wiegreffe-pinter-2019-attention, zhang2022greaselm, yasunaga2021qa}. Recent approaches like GraphMask attempt to improve faithfulness in explanations, but challenges persist in quantifying the fidelity of graph encoder explanations in LM-KG models \cite{schlichtkrull2021interpreting, aglionby2022faithful}.

\section{Model Architecture}

\subsection{Knowledge Graph Enhanced Commonsense Reasoning} 
\label{sec: model}

In this study, we focus on a category of models that synergise a text encoder (LM) and a knowledge graph encoder for the purpose of commonsense question answering. These models effectively combine linguistic and structured world knowledge to enhance reasoning and understanding. In a multi-choice commonsense question answering setting, the model processes a question $q$ and a set of answer choices $\mathcal{C}$. For each answer choice $a \in \mathcal{C}$, a concatenated input statement $S = [\boldsymbol{q}; \boldsymbol{a}]$ is formed, where $\boldsymbol{q}$ and $\boldsymbol{a}$ denote the vector representations of question
and option. The external Knowledge Graph is then utilized to extract a relevant subgraph $\mathcal{G}$, guided by the input statement $S$. This contextualized subgraph is formally defined as a multi-relational graph $\mathcal{G} = (\mathcal{V}, \mathcal{I}, \phi)$, where $\mathcal{V}$ represents the set of vertices (or nodes), $\mathcal{I}$ the set of edges, and $\phi$ the relational types in the graph.
The language model, denoted as $\mathrm{LM}$, computes the context embedding $\boldsymbol{z} = \mathrm{LM}(S)$. This involves encoding the concatenated question and answer choice into a high-dimensional vector space, capturing the linguistic nuances and semantic relationships. 

Simultaneously, a graph encoder $f_\mathrm{G}$ is employed to encode the KG subgraph $\mathcal{G}$. The encoding $\boldsymbol{g} = f_\mathrm{G}(\mathcal{G})$ captures the structured relational information and knowledge present in the graph. Finally, a fusion module $F$ integrates the outputs of both the $\mathrm{LM}$ and $f_\mathrm{G}$ encoders to generate a joint representation $F(\boldsymbol{z}, \boldsymbol{g})$. 
This module can range from simple feature concatenation to more complex architectures, such as a transformer-based fusion layer, which effectively merges the linguistic context with the structured knowledge graph information. The output of this fusion model is then utilized to predict the plausible answer
$\mathrm{Y}$ from the set of choices. The joint representation $F(\boldsymbol{z}, \boldsymbol{g}))$ is then passed through a Multilayer Perceptron (MLP) to generate the final prediction from the set of choices $\mathcal{C}$. Formally, the training and prediction $\rho(\boldsymbol{q}, \boldsymbol{a})$ can be represented as:

\begin{equation}
\begin{aligned}
\mathrm{Y} = \rho(\boldsymbol{q}, \boldsymbol{a}) = \operatorname{argmax}_{a \in \mathcal{C}} \operatorname{MLP}(F(\boldsymbol{z}, \boldsymbol{g}))) \\
\text{s.t. }
\mathbb{L}=\mathbb{E}_{\boldsymbol{q}, \hat{\boldsymbol{a}}, \mathcal{C}}\left[-\log \frac{\exp (\rho(\boldsymbol{q}, \hat{\boldsymbol{a}}))}{\sum_{\boldsymbol{a} \in \mathcal{C}} \exp (\rho(\boldsymbol{q}, \boldsymbol{a}))}\right]
\end{aligned}
\end{equation}

where $\operatorname{argmax}$ selects the answer choice $a$ that maximises the output of the MLP applied to the joint representation. During training, we maximise the plausibility score of the correct answer $\hat{\boldsymbol{a}}$ by minimising the cross-entropy loss. We give detail of KG encoding ($f_\mathrm{G}(\mathcal{G})$) in Appendix \ref{gnn}

\subsection{Post-hoc LM-KG Explanation Framework}

Perturbation-based methods are often used to provide instance-level explanations. In this context, perturbations are derived by sequentially masking out the most weighted groups of connected edges in the knowledge graph, focusing specifically on the most weighted path connecting context nodes and the predicted answer node. 


Given a graph $\mathcal{G} = (\mathcal{V}, \mathcal{I}, \phi)$, where nodes are represented by an attribute matrix $T \in \mathbb{R}^{n \times d}$ and edges by an adjacency matrix $A \in \mathbb{R}^{n \times n}$. The goal of post-hoc explanation is to identify a subgraph $\mathcal{G}^{\prime}$ with binary importance masks $M_A \in [0,1]^{n \times n}$ on the adjacency matrix and $M_T \in [0,1]^{n \times d}$ on the node attributes, respectively. Formally, the subgraph is defined as $\mathcal{G}^{\prime} = \{A \odot M_A ; T \odot M_T\}$, where $\odot$ denotes elementwise multiplication. 

Following the Feature Removal Principle \cite{covert2020feature}, when ground-truth explanations are not available, we assess the explanation's effectiveness by measuring the model's sensitivity to explanations $\mathcal{G}^{\prime}$. This could be done by sequentially masking out the most critical sets of nodes indicated by $M_A$ that follows edge attention weights \(\alpha\) and observing the drop in performance \cite{yuan2022explainability}. This approach ensures that the most important nodes are recognised by the rate at which the model's accuracy deteriorates when these nodes are not functioning.


Mathematically, the degradation is defined as:
\begin{equation}
\begin{aligned}
\Delta \text{Acc}(\hat{T}_n) &= f_\mathrm{G}(\mathcal{G}) - f_\mathrm{G}(\mathcal{G}^{\prime}) \\
\text{s.t.  } \mathcal{G}^{\prime} &\sim \mathcal{B}(\mathcal{G}, \alpha, A, n, T)
\end{aligned}
\end{equation}
where $\hat{T}_n$ denotes the set of $n$ most influential nodes. $\mathcal{B}$ represents the perturbations applied to the original node attribute matrix ${T}$. $\Delta \text{Acc}$ quantifies the rate at which the accuracy decreases when pruning is applied. 

\begin{figure}[t]
\centering
\includegraphics[width=0.8\columnwidth]{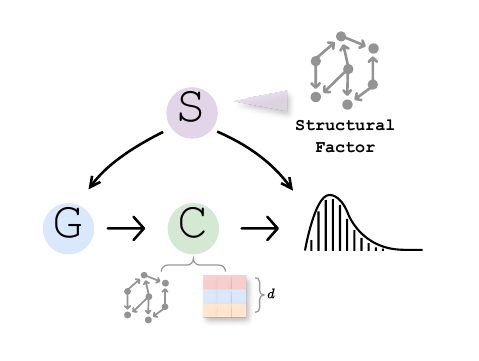} 
\caption{Behavior of GNN model from the causality perspective in the form of Structural Equation Model. There are two possible causal paths can be found.}
\label{sem}
\end{figure}

\section{Spurious GNN Causality}

Inspired by \citet{zhao2023towards}, spurious explanations refer to those that do not align with the true reasoning behind the predictions on \( \mathcal{G} \), rendering the extraction of \( \mathcal{G}^{\prime} \) for explanations anecdotal. To illustrate this, we can model the GNN using a Structural Equation Model (SEM) as depicted in Figure \ref{sem}. Here, variable \( C \) represents discriminative causal factors, while variable \( S \) denotes confounding environmental factors. The GNN learned from \( f_{G} \) might maintain prediction distribution \( Y \) due to the confounding effects of distribution shifts or differing causal variables from the original \( \mathcal{G} \). This issue is exacerbated in weakly-trained unstable GNNs in LM-KG models, making GNNs predictions unreliable. The model's inference process can be broken down into two paths:

\begin{enumerate}
    \item \( \mathcal{G} \rightarrow C \rightarrow Y \): The causal path lies behind the inference process, with the representation encoding the critical variables \( C \). This path utilises information from the entire input graph \( \mathcal{G} \).
    \item \( \mathcal{G} \leftarrow S \rightarrow Y \): The confounding effect of the spurious factor \( S \) can influence the inference process by leading the model to neglect the semantics of node embeddings. Especially when an input graph \( \mathcal{G}^{\prime} \) is out-of-distribution (OOD), the supportive GNN may fail to reflect its discriminative features. During inference, the encoded representation of \( \mathcal{G} \) is distant from those seen in the training set, making the generalise unreliably. This effect will be transferred through fusion layers to the LM, leading to better accuracy but unreliable explanations.
\end{enumerate}

To gain a deeper understanding of the reasons behind this problem, we can examine the behavior of a state-of-the-art LM-KG model from a causality perspective. The GSC \cite{wang2021gnn} model provides a clear illustration of this issue. They use Sparse-VD \cite{molchanov2017variational} to analyse GNN components in many LM-KG commonsense QA systems and find that the \textit{counting of edges} in the knowledge graph plays a crucial role in the reasoning process for LM-KG models. Even a simple hard counting algorithm that counts the occurrence of each possible edge triplet can achieve QA performance comparable to complex GNN methods, but the attention mechanism and node embedding features in GNNs are not predominant. In such cases, especially when there is support of reasoning from the LM and the training data is relatively scarce, the message-passing process might fail to capture effective causal factors other than graph motifs, leading to the loss of significant symbolic nodes' ability, which are essential in the knowledge graph, thus ignoring essential causal relationships.

Addressing this issue requires careful consideration of the model's learning objective and the development of methods that can faithfully capture the causal factors contributing to the predictions.
\section{LM-KG Explanation Evaluation Metrics}
Here we evalute GNN explanability in a fusion model in two folds, namely, faithfulness and sparsity. With “faithful graph encoders”, we refer to GNN representations being able to reflect the genuine rationale of the prediction. While sparsity means rationales should capture the most important input features and ignore the irrelevant ones. We argue that LM-KG fusion models are intrinsically unable to provide graph-structured explanations that are highly faithful to the full model. 

\begin{figure}[t]
\centering
\includegraphics[width=1.05\columnwidth]{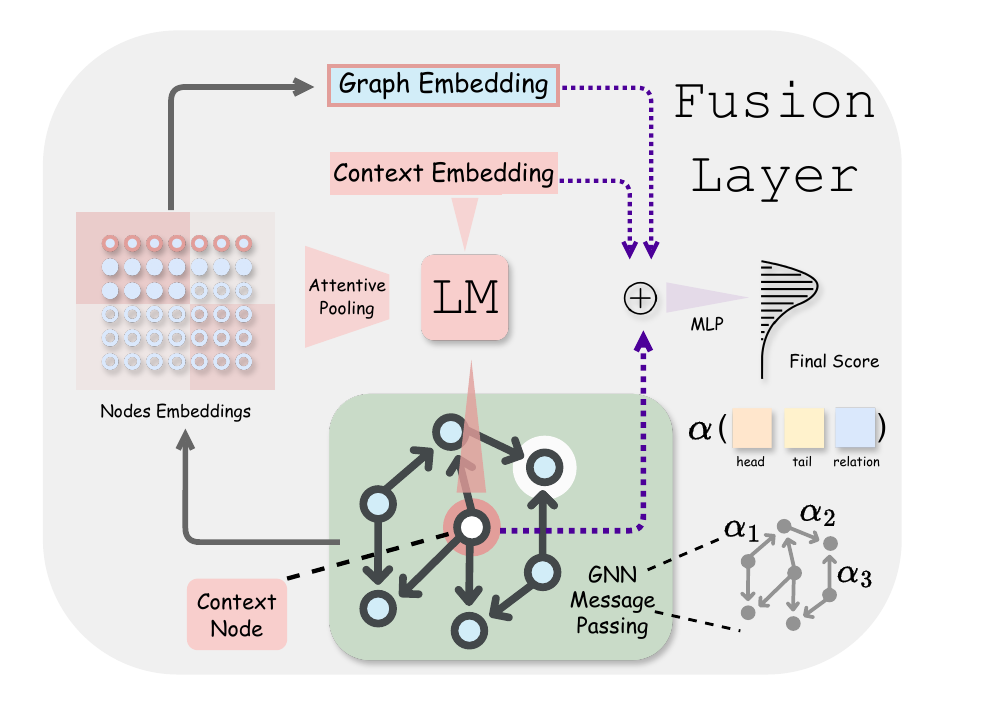} 
\caption{
This figure depicts the comprehensive structure of the fusion layer, through which the LM is deeply integrated with the KG. The components highlighted in \textcolor{pink}{pink} signify the modules that exhibit a strong correlation with the LM. The \textcolor{purple}{purple} dashed line denotes the specific segments that require LM detachment before the final prediction to keep GNN faithfulness.}
\label{archi}
\end{figure}

\subsection{LM-KG Fidelity}
Intuitively, If trustworthy explanations are to be extracted from the GNN, the GNN itself must demonstrate predominant reasoning ability within the overall model. Only then will the explanations extracted from the GNN be faithful and truly representative of the reasoning process. Hence, LM-KG Fidelity here is defined as the intersection of prediction between the original and the GNN with fundamental changes. Concretely, we define LM-KG Fidelity ($\mathrm{F}_\mathrm{KG}$) as the prediction agreement between the original model and the language model factors detached model output. 

\subsubsection{Proxy for Faithfulness} 
To maintain isolation and integrity of the GNN model, we steadily prune text encoder from the fusion layer without further training, as shown in Figure \ref{archi}. Inspired by \cite{schlichtkrull2021interpreting, aglionby2022faithful}, $\mathrm{F}_\mathrm{KG}$ is conducted using a controlled variable method with complementary masking, all factors are kept constant except that the text encoder reasoning components are totally detached from the interaction between modalities in the fusion layer. Keeping nodes features and the model architecture as is allows us to establish a causal relationship between the text encoder variable and the observed outcomes, especially in such a model class with multiple deep fusions. Pruning here can be equivalently thought of as adding a certain type of noise when prediction, it contains at best minimal useful information for answering the question correctly. It can be categorised as belonging to the class of perturbation-based methods \cite{guan2019towards,schlichtkrull2021interpreting}.

Specifically, follow \citet{wang2022reinforced} $\mathrm{F}_\mathrm{KG}$ is defined as:

\begin{equation}
\small
\begin{aligned}
\mathrm{F}_\mathrm{KG} &= \frac{\mathrm{d}_\mathrm{H}(\hat{\mathrm{C}}_{\mathcal{M}}, \hat{\mathrm{C}}_{\mathcal{M}_{\setminus z}})}{N} \\
&= \frac{\sum_{i=1}^{N} \mathbb{I}\left(\hat{\mathrm{C}}_{\mathcal{M}}^{(i)}, \hat{\mathrm{C}}_{\mathcal{M}_{\setminus z}}^{(i)}\right)}{N} \\
\text{s.t. }
\mathrm{\hat{C}}_{\mathcal{M}} &= \underset{c \in C}{\arg \max} \, P\left( c \mid \mathcal{G}, \mathcal{M}\right), \\
\mathrm{\hat{C}}_{\mathcal{M}_{\setminus z}} &= \underset{c \in C}{\arg \max } \, P( c \mid \mathcal{G}, \mathcal{M}_{\setminus \boldsymbol{z}})
\end{aligned}
\end{equation}

The $\mathrm{F}_\mathrm{KG}$ score is defined as the normalised Hamming Distance $\mathrm{d}_\mathrm{H}$ which represents the proportion of instances where the predictions of the two models agree Where $C$ is the set of choices, \(\hat{\mathrm{C}}_{\mathcal{M}}^{(i)}\) and \(\hat{\mathrm{C}}_{\mathcal{M}_{\setminus z}^{(i)}}\) are the predictions for the \(i\)-th instance made by the original model and the complementary mask applied model \(\mathcal{M}_{\setminus \boldsymbol{z}}\) respectively. $P(c \mid \mathcal{M})$ denotes the probability distribution of the output $Y$ given the model $\mathcal{M}$. $\mathbb{I}(x, y)$ is the indicator function, which is 1 if $x = y$ and 0 otherwise. $N$ is the total number of instances in the dataset considered. Accuracy performance and comparison between the complete model's output and the LM-detached model's prediction are provided in the Figure \ref{fig:commonsenseqa_results} and \ref{fig:openbookqa_results}. Measurement of $\mathrm{F}_\mathrm{KG}$ is reported in Table \ref{tab: ori graph fid}.

\subsubsection{Fidelity of Consistency}
Note that the $\mathrm{F}_\mathrm{KG}$ metric studies the change of prediction accuracy. In order to quantitatively assess the divergence between the output density of our original model $\mathcal{M}$ and its pruned variant $\mathcal{M}_{\setminus z}$, we first devise the \textbf{LM-KG Consistency} ($\mathrm{C}_\mathrm{LK}$) metric to measure the alignment between the probability distributions of their outputs. Our chosen metric is inspired by the Jensen–Shannon divergence $J$ \cite{lin1991divergence}, a symmetrised and smoothed version of the Kullback-Leibler divergence \cite{kullback1951information}, which offers a bounded measure of similarity between probability distribution pairs. The $\mathrm{C}_\mathrm{LK}$ metric is computed as follows:
\begin{equation}
\small
\begin{aligned}
\mathrm{C}_\mathrm{LK}: J\left(\mathcal{M}, \mathcal{M}_{\setminus \boldsymbol{z}}\right) = \lambda &\mathbb{D}_{KL}\left(P\left(Y\mid\mathcal{M}\right) \| \mathcal{A}\right) \\
 +(1-\lambda) &\mathbb{D}_{KL}(P(Y\mid\mathcal{M}_{\setminus \boldsymbol{z}}) \| \mathcal{A})
\end{aligned}
\end{equation}

Where $\mathbb{D}_\mathrm{KL}$ represents the Kullback-Leibler divergence. The key to the computation of $J$ is the average of the two distributions.
$\mathcal{A}$ serves as the mid-point reference distribution against which the divergence of each of the two distributions is measured.
By employing $\mathrm{C}_\mathrm{LK}$ as our metric, we aim to capture the nuanced differences between the output probability distributions of $\mathcal{M}$ and $\mathcal{M}_{\setminus \boldsymbol{z}}$. A smaller $\mathrm{C}_\mathrm{LK}$ indicates a high degree of similarity or consistency between the two models, while a larger value signifies a greater divergence in their outputs, and even when the LM output is detached, the graph encoder can still assign probabilities to choices that closely align with the original model's decisions, making it potentially more representative of the original model's thought process. Note that $\mathrm{C}_\mathrm{LK}$ is more sensitive than $\mathrm{F}_\mathrm{KG}$. 

\subsection{LM-KG Explanation Sparsity}
Good explanations should be sparse, which means they should capture the most important input features and ignore the irrelevant ones. The metric Sparsity measures such a property. Specifically, it measures the fraction of features in the final GNN layer selected as important by explanation methods. Formally, we define it as the percent of important node embeddings masked in $T$. Note that we need to compare model explanation performance by combining sparsity with other criteria. Note that we must evaluate model explanation performance by jointly considering sparsity and other criteria. For models undergoing the same change in sparsity, those exhibiting greater performance variation indicate that the factors driving this change possess stronger explanatory power for the model.

\section{Methodology}
To achieve a more faithful LM-KG interpretation, it's imperative to ensure that the introduced modifications of models do not substantially deviate from the LM's behaviour, implying that after introducing modifications, the GNN encoder should predict a target distribution that mirrors the one emitted by the unaltered model to retain its subtle reasoning ability. While traditional methods have relied heavily on cross-entropy as the primary objective, the unfaithful GNN encoder of existing LM-KG models demands a more nuanced regularisation of training procedure. We next introduce \textbf{LM-KG Distribution-aware Alignment} (LKDA) to bridge this gap.

\subsection{Knowledge Graph Anchored Alignment through Divergence} 

\begin{figure}[t]
    \centering
    \begin{subfigure}{0.45\columnwidth}
        \centering
        \includegraphics[width=\textwidth]{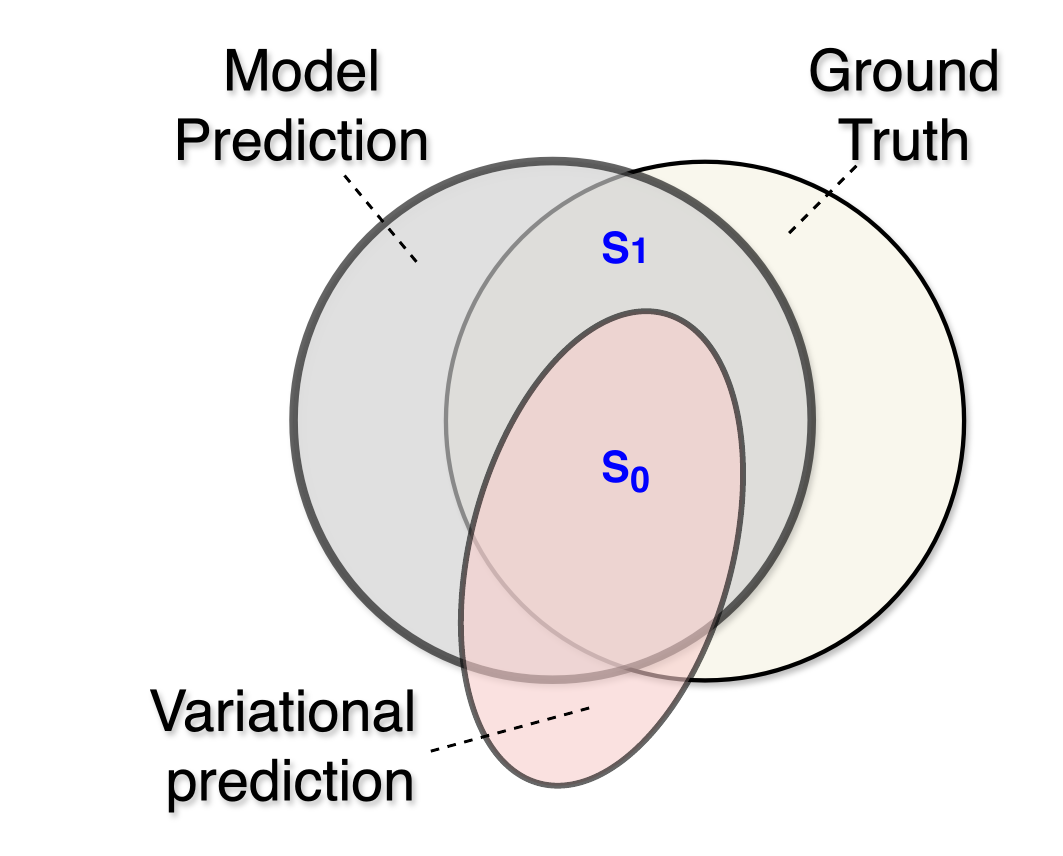}
        \caption{Previous Objective}
    \end{subfigure}
    \begin{subfigure}{0.45\columnwidth}
        \centering
        \includegraphics[width=\textwidth]{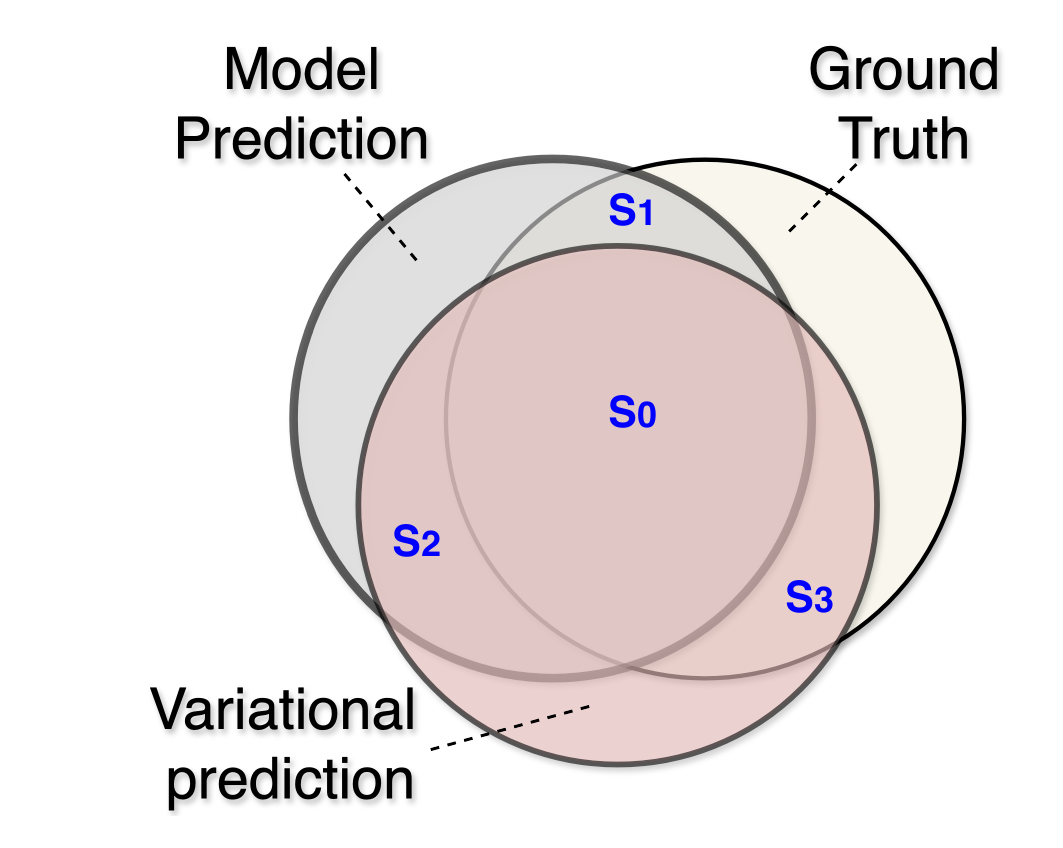}
        \caption{LKDA Objective}
    \end{subfigure}

    \caption{Illustration of our proposed new objective}
    \label{fig:overlap}
\end{figure}

\textbf{LKDA} enhances the cross-entropy $\mathbb{L}_{CE}$ by introducing a consistency regularisation $\mathbb{L}_{{C}_\mathrm{TG}}$. This factor is an alignment loss used as an auxiliary task incorporated into that ensures the graph encoder's target prediction align closely with the original model's predictions. \textbf{LKDA} is given by:

\begin{equation}
\small
\begin{aligned}
\mathbb{L}_\mathrm{align}(\hat{\mathrm{C}}_{\mathcal{M}}, \hat{\mathrm{C}}_{\mathcal{M}_{\setminus \boldsymbol{z}}}) = \nabla_{\theta_t} \mathbb{L}_{CE} + \lambda \cdot \nabla_{\theta_t} \mathbb{L}_{{C}_\mathrm{TG}} 
\end{aligned}
\end{equation}
\\
In this equation, $\theta_t$ are the model parameters at time step $t$, $\lambda$  controls the balance between prediction preservation and alignmen, $\mathbb{L}_{CE}$ represents the cross-entropy loss, which was traditionally employed. $\mathbb{L}_{{C}_\mathrm{TG}}$ is the consistency term that measures the divergence between the probability distributions of the original and LM-detached models. The equation shows the parameter update rule, where the gradients of the two losses are subtracted from the current parameters $\theta_t$ to obtain the updated parameters $\theta_{t+1}$. The algorithm details of this strategy can be found in Appendix \ref{LKDA algo}.

\subsection{Theoretical Analysis}

From our previous discussions, it is evident that $\mathcal{G}^{\prime}$ obtained via Equation 1 cannot be reliably used as explanations. One critical issue with existing GNN explanation methods lies in the inductive bias: achieving the same outcomes does not guarantee the same underlying causes, leaving these approaches vulnerable to spurious explanations. This is illustrated in Figure \ref{fig:overlap}. The objective proposed in Equation 1 optimizes the mutual information between the model prediction $Y$ and the ground truth $T$, which corresponds to maximizing the overlap $S_0 \cup S_1 $ between $I(T; Y)$ in Figure \ref{fig:overlap}(a) and Figure \ref{fig:overlap}(b).

However, this learning target cannot prevent the generation of spurious explanations. Provided $KG$ explanation may fall into region $S_1 \cup S_3$, which does not faithfully represent model reasoning. Instead, a more sensible objective should be maximizing region $S_0 \cup S_2 $ in Figure \ref{fig:overlap}(b). The intuition behind this is that in the search input space that causes the same outcome, no matter correct or wrong, the identified $\mathcal{G}^{\prime}$ should account for both the representative and discriminative parts of the original LM-KG model, to prevent both unfaithful KG and spurious explanations that produce the same outcomes due to different causes. Ensuring the alignment of \( \mathcal{M} \) and \( \mathcal{M}_{\setminus \boldsymbol{z}} \) while increasing the area of \( S_0 \) will inevitably reduce the area of \( S_3 \cup S_4 \). Therefore, our method can reduce the occurrence of incorrect or shortcut spurious explanations.

\begin{table}[t]
\centering
\small
\begin{tabular}{ccc}
\toprule
\textbf{Method} & IH-dev (\%) & IH-test (\%) \\
\midrule
QA-GNN  & 76.1 & 73.3 \\
\rowcolor{gray!25} +\textit{LKDA} & 76.3$_{\uparrow0.2}$ & 73.4$_{\uparrow0.1}$ \\
\midrule
GreaseLM  & 77.4 & \textbf{74.2} \\
\rowcolor{gray!25} +\textit{LKDA} & \textbf{77.8}$_{\uparrow0.4}$ & \textbf{74.2}$_{\uparrow0.0}$ \\
\midrule
MHGRN  & 74.4 & 71.1 \\
\rowcolor{gray!25} +\textit{LKDA} & 76.9$_{\uparrow2.5}$ & 71.2$_{\uparrow0.1}$ \\
\bottomrule
\end{tabular}
\caption{Accuracy comparison of three different LM-KG models in their original version and trained with the LKDA scheme (grey background) on the CommonsenseQA dataset.}
\label{mask_result_commonsenseqa}
\end{table}

\begin{table}[t]
\centering
\small
\begin{tabular}{ccc}
\toprule
\textbf{Method} & Dev (\%) & Test (\%) \\
\midrule
QA-GNN  & 72.4 & 70.4 \\
\rowcolor{gray!25} +\textit{LKDA} & 79.0$_{\uparrow6.6}$ & 80.0$_{\uparrow9.6}$ \\
\midrule
GreaseLM & 73.4 & 71.6 \\
\rowcolor{gray!25} +\textit{LKDA} & \textbf{80.6}$_{\uparrow7.2}$ & \textbf{82.4}$_{\uparrow10.8}$ \\
\midrule
MHGRN  & 69.4 & 67.4 \\
\rowcolor{gray!25} +\textit{LKDA} & 71.2$_{\uparrow1.8}$ & 66.6$_{\downarrow0.8}$ \\
\bottomrule
\end{tabular}
\caption{Accuracy comparison of three different LM-KG models in their original version and trained with the LKDA scheme (grey background) on the OpenBookQA dataset.}
\label{mask_result_openbookqa}
\end{table}

\section{Experiment Settings}
\begin{figure}[t]
    \centering
    \begin{subfigure}{0.45\columnwidth}
        \centering
        \includegraphics[width=\textwidth]{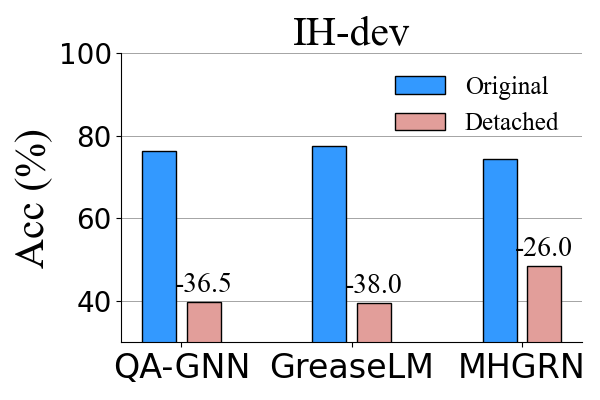}
        \caption{}
    \end{subfigure}
    \begin{subfigure}{0.45\columnwidth}
        \centering
        \includegraphics[width=\textwidth]{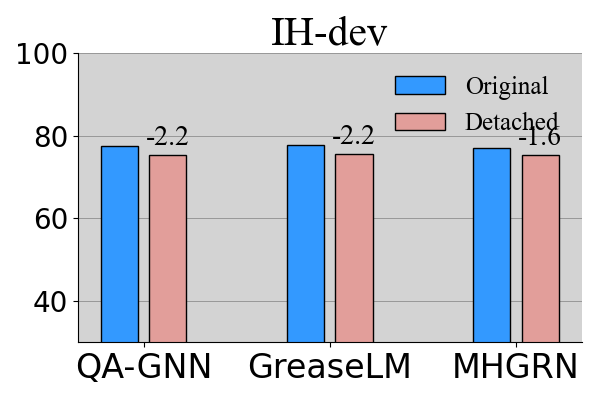}
        \caption{}
    \end{subfigure}
    \begin{subfigure}{0.45\columnwidth}
        \centering
        \includegraphics[width=\textwidth]{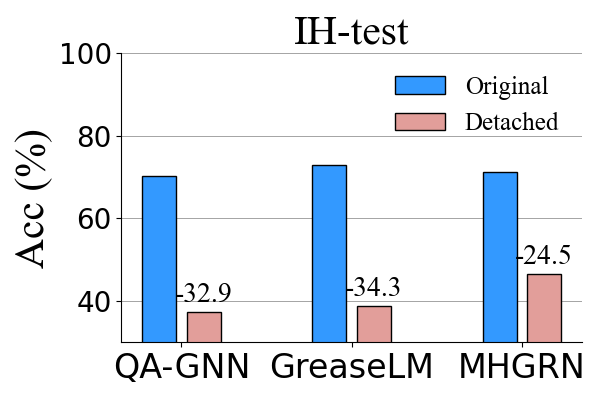}
        \caption{}
    \end{subfigure}
    \begin{subfigure}{0.45\columnwidth}
        \centering
        \includegraphics[width=\textwidth]{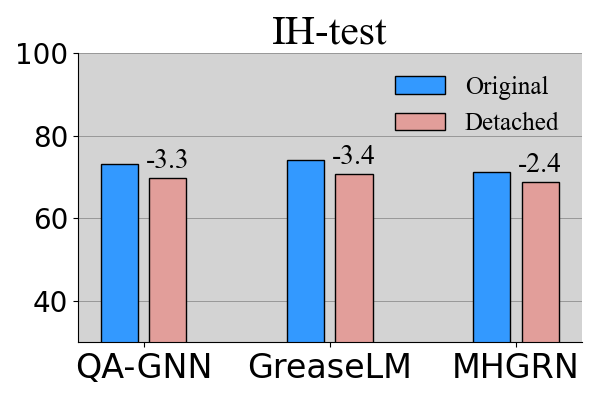}
        \caption{}
    \end{subfigure}
    \caption{The bar charts compare the accuracy of the model on CommonsenseQA before and after LKDA training when the LM is detached. The models trained with LKDA are shown with a gray background.}
    \label{fig:commonsenseqa_results}
\end{figure}

\subsection{Datasets \& KG}
We assess our methods using two multiple-choice question answering datasets: CommonsenseQA \textbf{in-house (IH)} data split \cite{talmor2019commonsenseqa, lin2019kagnet} and OpenBookQA \cite{mihaylov2018can}, serving as benchmarks for commonsense reasoning. We also use ConceptNet \cite{speer2017conceptnet}, a broad commonsense KG, for our tasks. Details can be found in Appendix \ref{data}.

\subsection{LM-KG Faithfulness Baseline Models}
To assess our \textit{LKDA} training and LM-KG Fidelity metric, we compare it with three LM-KG models: QA-GNN \cite{yasunaga2021qa}, GreaseLM \cite{zhang2022greaselm}, and MHGRN \cite{feng2020scalable}. Each uniquely integrates language models with knowledge graphs: QA-GNN uses a context node, GreaseLM enhances interaction through a fusion mechanism, and MHGRN offers a multi-hop relational reasoning architecture.

For fair comparison, we use RoBERTa-Large \cite{liu2019roberta} model and its generated concepts embedding for our experiments.

We also include the $\mathrm{Train TE}$ \cite{aglionby2022faithful} ($-TE$) ablation for faithfulness comparison, freezing text encoder weights to enhance the GNN's reasoning contribution. Unlike the $\mathrm{-Embed}$ \cite{aglionby2022faithful} ablation, which detaches the text encoder only from the final $\mathrm{MLP}$, $-TE$ better aligns with our goal. Implementation and hyper-parameters are detailed in Appendix \ref{hyper}.
\begin{table}[t]
\centering
\tiny
\begin{tabular}{llllll}
\toprule
         & \multirow{2}{*}{Model} & \multicolumn{2}{l}{\textbf{CommonsenseQA}} & \multicolumn{2}{l}{\textbf{OpenBookQA}} \\
         & & IH-dev        & IH-test        & dev        & test        \\ \hline \noalign{\vspace{0.6ex}}
\multirow{2}{*}{\begin{sideways}-TE\end{sideways}} & QA-GNN$_{TE}$    & 33.5   & 30.5  & 45.6  & 45.5      \\
         & MHGRN$_{TE}$ & 29.7       & 24.5       & 44.8      & 41.0       \\
        \hline\hline  \noalign{\vspace{0.6ex}}
\multirow{3}{*}{\begin{sideways}$\mathcal{M}_{\setminus \boldsymbol{z}}$\end{sideways}} & QA-GNN    & 43.5   & 39.8  & 39.3  & 45.5      \\
         & GreaseLM & 41.2       & 40.7       & 60.3      & 62.7        \\
        & MHGRN    & 52.3       & 51.0       & 75.4        & 73.0      \\
        \bottomrule \noalign{\vspace{0.6ex}}
\multirow{3}{*}{\begin{sideways}LKDA\end{sideways}}&QA-GNN           & 98.5   & \textbf{98.7}$_{\uparrow58.9}$  & 97.6  & 98.0  \\
&GreaseLM         & \textbf{98.9}$_{\uparrow57.7}$     & 98.0    & \textbf{99.6}$_{\uparrow39.6}$  & \textbf{99.6}$_{\uparrow36.9}$   \\
&MHGRN              & 95.5 & 95.0  & 96.2  & 97.4 \\
\bottomrule
\end{tabular}
\caption{LM-KG Fidelity measurement of three LM-KG models variations on two datasets.}
\label{tab: ori graph fid}
\end{table}

\section{Results Analysis \& Discussion}
Table \ref{tab: ori graph fid} presents the \textbf{LM-KG Fidelity} results on CommonsenseQA and OpenBookQA for LKDA-trained models and three LM fully detached models. LKDA notably enhances faithfulness across all scenarios, with GreaseLM$_{LKDA}$ on the CommonsenseQA IH-dev split achieving a 57.7\% and QAGNN$_{LKDA}$ on the IH-test split achieving a 58.9\% accuracy increase. This highlights LKDA's effectiveness in addressing model unfaithfulness and bolstering graph encoder predictions, thus laying a foundation for reliable graph interpretation. Additionally, Tables \ref{mask_result_commonsenseqa} and \ref{mask_result_openbookqa} report accuracy under original models and LKDA settings. It is noteworthy that these tables show consistent improvements, including an over 11\% improvement for GreaseLM on the OpenBookQA test dataset.

\subsection{LM-detached Models}
\begin{figure}[t]
    \centering
    \begin{subfigure}{0.45\columnwidth}
        \centering
        \includegraphics[width=\textwidth]{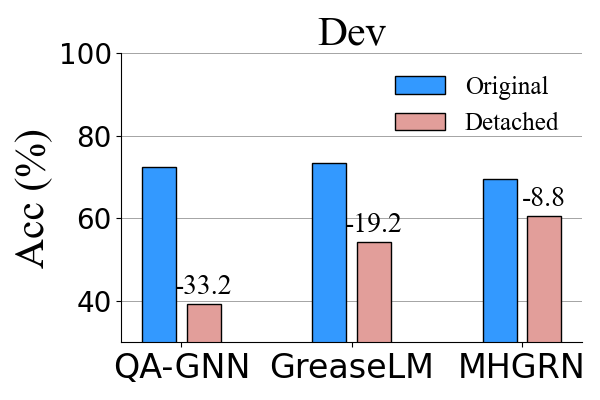}
        \caption{}
    \end{subfigure}
    \begin{subfigure}{0.45\columnwidth}
        \centering
        \includegraphics[width=\textwidth]{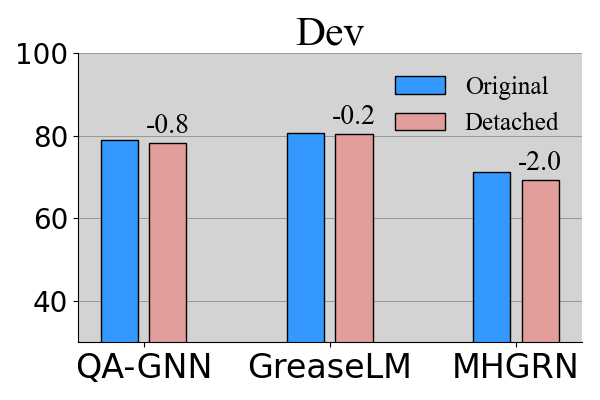}
        \caption{}
    \end{subfigure}
    \begin{subfigure}{0.45\columnwidth}
        \centering
        \includegraphics[width=\textwidth]{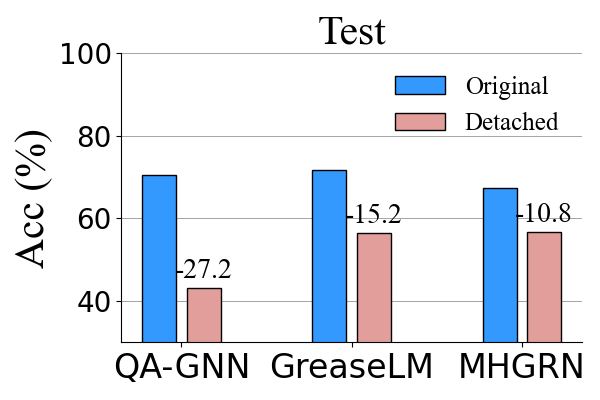}
        \caption{}
    \end{subfigure}
    \begin{subfigure}{0.45\columnwidth}
        \centering
        \includegraphics[width=\textwidth]{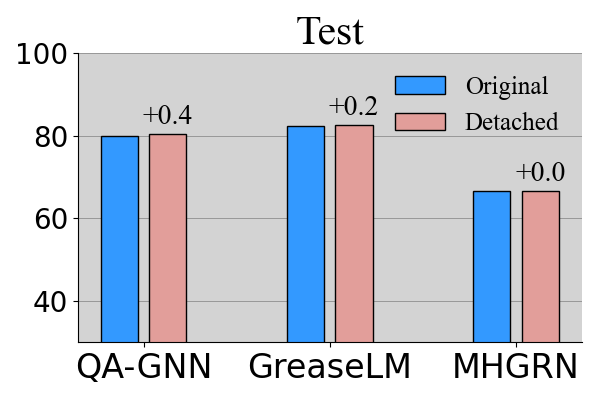}
        \caption{}
    \end{subfigure}
    \caption{The bar charts compare the accuracy of the model on OpenBookQA before and after LKDA training when the LM is detached. The models trained with LKDA are shown with a gray background.}
    \label{fig:openbookqa_results}
\end{figure}

Figures \ref{fig:commonsenseqa_results} and \ref{fig:openbookqa_results} show that removing the text encoder significantly drops performance in all models. For instance, in CommonsenseQA IH-dev, GreaseLM's accuracy drops by 39.7\%. This highlights the text encoder's crucial role. However, LKDA models without the LM embedding show only minor drops or slight improvements in accuracy. This suggests the graph encoder now has the most influence, ideal for reliable explanations.

LKDA-trained models consistently outperform those without fidelity regularization. On the OpenBookQA test set, QA-GNN$_{LKDA}$ achieves 80.0\% accuracy, a 9.6\% increase over the vanilla QA-GNN. GreaseLM$_{LKDA}$ achieves 82.4\%, surpassing the original by 10.8\%, and matches the fine-tuned T5 model. This indicates that LKDA improves reasoning in the graph encoder, making it a reliable proxy for the model's reasoning process.

\subsection{LM-KG Fidelity}
Table \ref{tab: ori graph fid} shows that $\mathrm{F}_\mathrm{KG}$ scores significantly increased after LKDA training. In CommonsenseQA IH-test, QA-GNN's fidelity rose from 75.7\% to 98.7\%, GreaseLM from 40.7\% to 98.0\%, and MHGRN from 51.0\% to 95.0\%. All models showed over 95\% $F_\mathrm{KG}$, indicating high faithfulness of graph encoders to the original model outputs. GreaseLM's fidelity improved notably, achieving 99.6\% on OpenBookQA dev and test sets, demonstrating LKDA's effectiveness.

\subsection{Explanation Fidelity}
Evaluating the explainability of the obtained GNNs is challenging due to the lack of commonsense KG explanation ground-truth. We specifically study this by observing prediction changes when sequentially removing important nodes from the final GNN layer. We define importance as the attention weights ($\alpha$ in Figure \ref{archi}) between the head node and tail nodes learned by the model to test its explanation performance. Generally, the removal of truly important edges would significantly degrade the classification performance. Thus, a faster performance drop represents stronger fidelity. 

\begin{figure}[t]
    \centering
    \begin{subfigure}[t]{0.45\columnwidth}
        \centering
        \includegraphics[width=\linewidth]{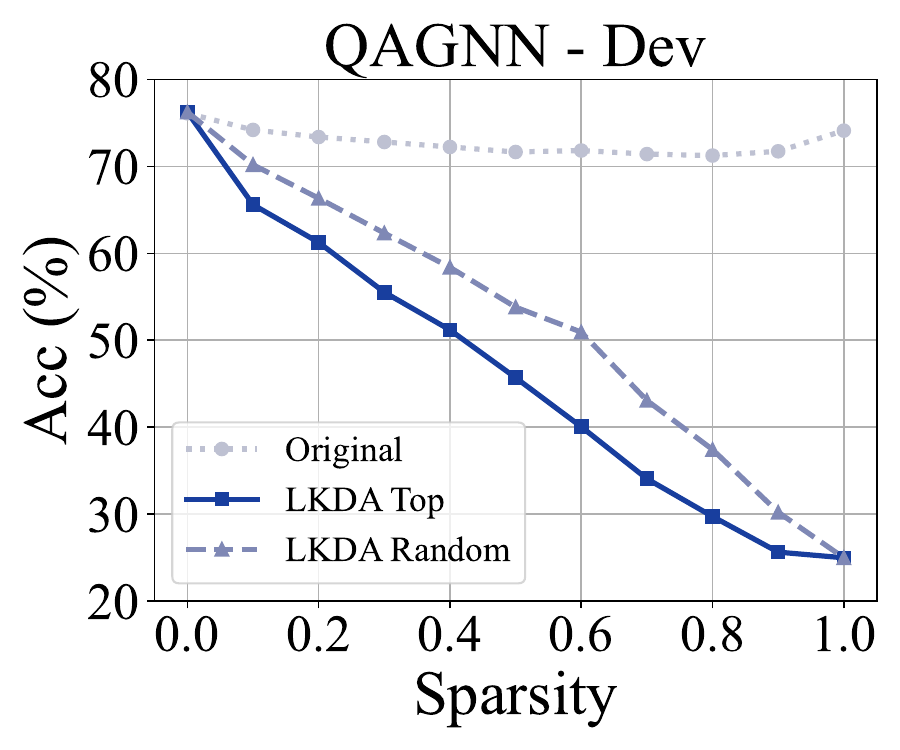}
        \caption{}
        \label{fig:figure1}
    \end{subfigure}
    \hfill
    \begin{subfigure}[t]{0.45\columnwidth}
        \centering
        \includegraphics[width=\linewidth]{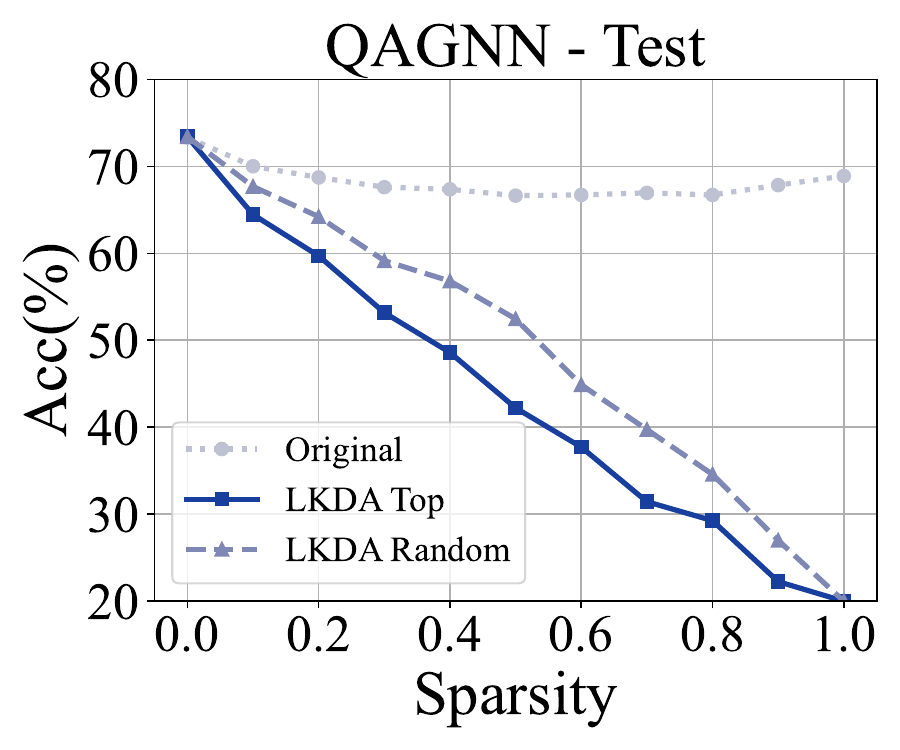}
        \caption{}
        \label{fig:figure2}
    \end{subfigure}
    \vskip\baselineskip
    \begin{subfigure}[t]{0.45\columnwidth}
        \centering
        \includegraphics[width=\linewidth]{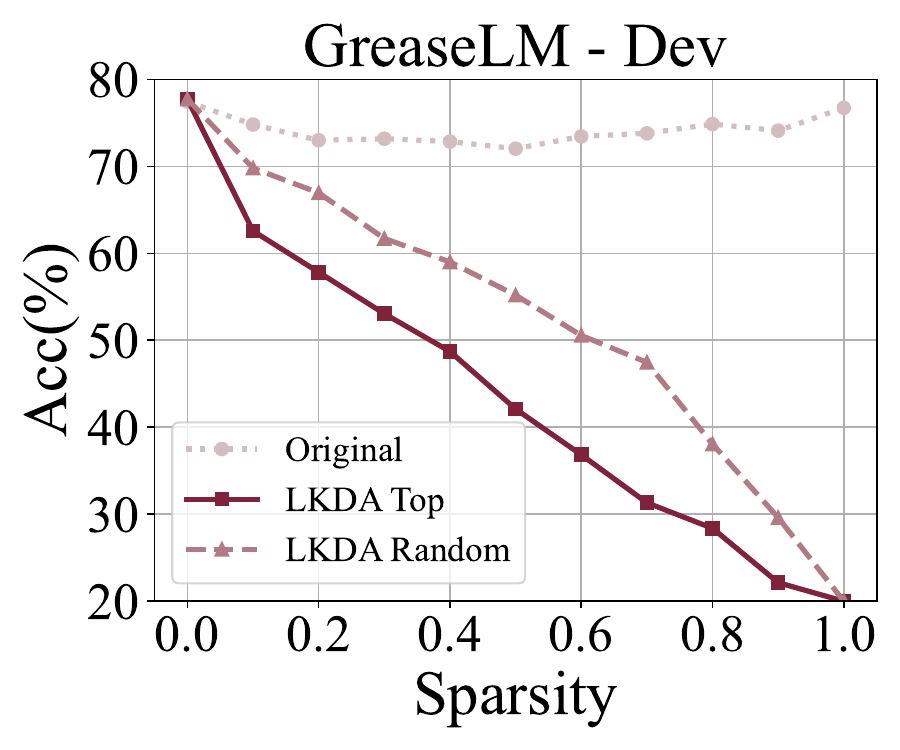}
        \caption{}
        \label{fig:figure3}
    \end{subfigure}
    \hfill
    \begin{subfigure}[t]{0.45\columnwidth}
        \centering
        \includegraphics[width=\linewidth]{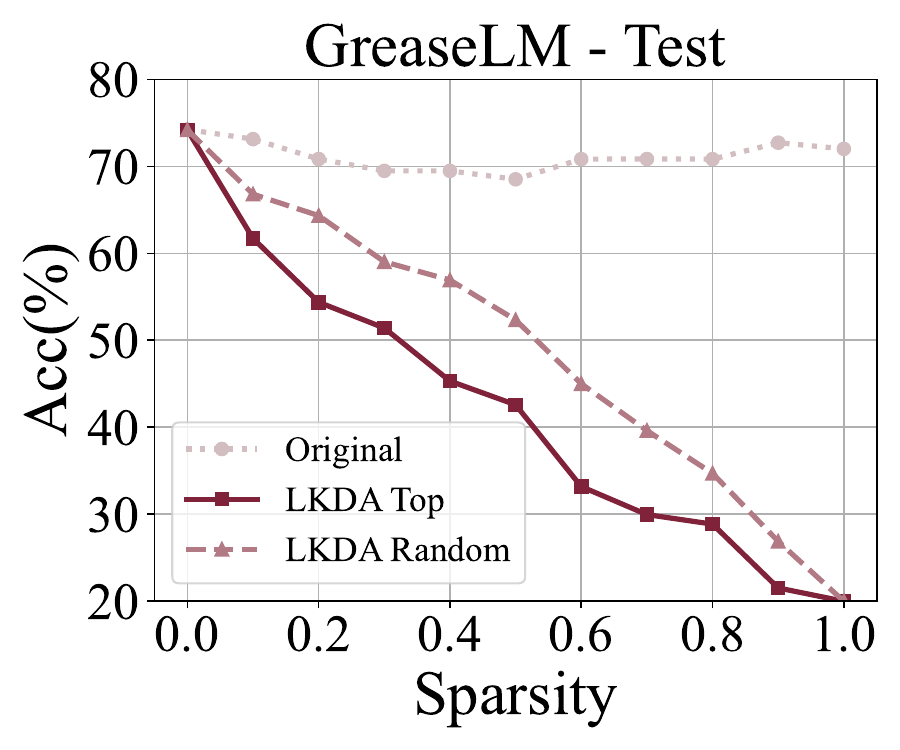}
        \caption{}
        \label{fig:figure4}
    \end{subfigure}
    \caption{The line graphs depict Fidelity-Sparsity results of three variants of QAGNN and GreaseLM on CommonsenseQA. Faster accuracy drops with increasing sparsity indicate stronger fidelity and more effective explanations.}
    \label{fig:overall}
\end{figure}

Figures \ref{fig:overall} show the results of comparing the explanability of original models and LKDA architectures of QAGNN and GreaseLM on CommonsenseQA. We analyse the effect on model target predictions by incrementally removing node features, thereby increasing sparsity, and jointly evaluating both sparsity and fidelity. The experiments are divided into three variants: 
\begin{itemize}
    \item Feature reduction on the original model (\textsc{Original})
    \item Random removal of node features on the LKDA-aligned model (\textsc{Random})
    \item Masking nodes according to the magnitude of edge attention values (\textsc{Top})
\end{itemize}

As shown in Figure \ref{fig:overall}, as GNN sparsity increases, both random and top methods exhibit a much more rapid accuracy drop compared to the original versions. For example, after sparsity increases to 0.1, the accuracy of the original QA-GNN remains relatively steady on both dev and test sets, while for LKDA, the accuracy drops by around 10\%, indicating that the explanations from LKDA better capture the critical edges. The more rapid degradation for LKDA as important edges are removed demonstrates that its explanations can better reflect the true reasoning process. Moreover, in all the figures, it is evident that at the same sparsity level, the accuracy drop of the top method is consistently faster than that of the random method. This observation further validates the effectiveness of the attention mechanism in identifying the most critical edges for the model's prediction. This analysis provides quantitative evidence that the knowledge graph explanations extracted from the LKDA model are more faithful and plausible.

\section{Limitations}
While \textit{LKDA} enhances explanation faithfulness in LM-KG models, some limitations exist. Evaluation relies on perturbation methods due to lack of ground-truth explanations, which may not fully capture explanation.
\textit{LKDA} introduces computational overhead, potentially restricting applicability to larger models and datasets.
\textit{LKDA} assumes a specific LM-KG architecture, and adapting it to other architectures may require further modifications. Quantitative metrics should be complemented with human evaluations to assess plausibility and understandability. Future research should incorporate user studies.

\bibliography{anthology,custom}

\begin{thebibliography}{36}
\expandafter\ifx\csname natexlab\endcsname\relax\def\natexlab#1{#1}\fi

\bibitem[{Aglionby and Teufel(2022)}]{aglionby2022faithful}
Guy Aglionby and Simone Teufel. 2022.
\newblock Faithful knowledge graph explanations in commonsense question answering.
\newblock In \emph{Proceedings of the 2022 Conference on Empirical Methods in Natural Language Processing}, pages 10811--10817.

\bibitem[{Covert et~al.(2020)Covert, Lundberg, and Lee}]{covert2020feature}
Ian Covert, Scott Lundberg, and Su-In Lee. 2020.
\newblock Feature removal is a unifying principle for model explanation methods.
\newblock \emph{arXiv preprint arXiv:2011.03623}.

\bibitem[{Danilevsky et~al.(2020)Danilevsky, Qian, Aharonov, Katsis, Kawas, and Sen}]{danilevsky2020survey}
Marina Danilevsky, Kun Qian, Ranit Aharonov, Yannis Katsis, Ban Kawas, and Prithviraj Sen. 2020.
\newblock A survey of the state of explainable ai for natural language processing.
\newblock In \emph{Proceedings of the 1st Conference of the Asia-Pacific Chapter of the Association for Computational Linguistics and the 10th International Joint Conference on Natural Language Processing}, pages 447--459.

\bibitem[{Feng et~al.(2020)Feng, Chen, Lin, Wang, Yan, and Ren}]{feng2020scalable}
Yanlin Feng, Xinyue Chen, Bill~Yuchen Lin, Peifeng Wang, Jun Yan, and Xiang Ren. 2020.
\newblock Scalable multi-hop relational reasoning for knowledge-aware question answering.
\newblock In \emph{Proceedings of the 2020 Conference on Empirical Methods in Natural Language Processing (EMNLP)}, pages 1295--1309.

\bibitem[{Guan et~al.(2019)Guan, Wang, Zhang, Chen, He, and Xie}]{guan2019towards}
Chaoyu Guan, Xiting Wang, Quanshi Zhang, Runjin Chen, Di~He, and Xing Xie. 2019.
\newblock Towards a deep and unified understanding of deep neural models in nlp.
\newblock In \emph{International conference on machine learning}, pages 2454--2463. PMLR.

\bibitem[{Herman(2017)}]{herman2017promise}
Bernease Herman. 2017.
\newblock The promise and peril of human evaluation for model interpretability.
\newblock \emph{arXiv preprint arXiv:1711.07414}.

\bibitem[{Hirschman and Gaizauskas(2001)}]{hirschman2001natural}
Lynette Hirschman and Robert Gaizauskas. 2001.
\newblock Natural language question answering: the view from here.
\newblock \emph{natural language engineering}, 7(4):275--300.

\bibitem[{Jain and Wallace(2019)}]{jain-wallace-2019-attention}
Sarthak Jain and Byron~C. Wallace. 2019.
\newblock \href {https://doi.org/10.18653/v1/N19-1357} {{A}ttention is not {E}xplanation}.
\newblock In \emph{Proceedings of the 2019 Conference of the North {A}merican Chapter of the Association for Computational Linguistics: Human Language Technologies, Volume 1 (Long and Short Papers)}, pages 3543--3556, Minneapolis, Minnesota. Association for Computational Linguistics.

\bibitem[{Kipf and Welling(2017)}]{kipf2017semisupervised}
Thomas~N. Kipf and Max Welling. 2017.
\newblock \href {https://openreview.net/forum?id=SJU4ayYgl} {Semi-supervised classification with graph convolutional networks}.
\newblock In \emph{International Conference on Learning Representations}.

\bibitem[{Kullback and Leibler(1951)}]{kullback1951information}
Solomon Kullback and Richard~A Leibler. 1951.
\newblock On information and sufficiency.
\newblock \emph{The annals of mathematical statistics}, 22(1):79--86.

\bibitem[{Li et~al.(2022)Li, Li, Shang, Dong, Sun, Liu, Ji, Jiang, and Liu}]{li2022pre}
Shaobo Li, Xiaoguang Li, Lifeng Shang, Zhenhua Dong, Cheng-Jie Sun, Bingquan Liu, Zhenzhou Ji, Xin Jiang, and Qun Liu. 2022.
\newblock How pre-trained language models capture factual knowledge? a causal-inspired analysis.
\newblock In \emph{Findings of the Association for Computational Linguistics: ACL 2022}, pages 1720--1732.

\bibitem[{Lin et~al.(2019)Lin, Chen, Chen, and Ren}]{lin2019kagnet}
Bill~Yuchen Lin, Xinyue Chen, Jamin Chen, and Xiang Ren. 2019.
\newblock Kagnet: Knowledge-aware graph networks for commonsense reasoning.
\newblock In \emph{Proceedings of the 2019 Conference on Empirical Methods in Natural Language Processing and the 9th International Joint Conference on Natural Language Processing (EMNLP-IJCNLP)}, pages 2829--2839.

\bibitem[{Lin(1991)}]{lin1991divergence}
Jianhua Lin. 1991.
\newblock Divergence measures based on the shannon entropy.
\newblock \emph{IEEE Transactions on Information theory}, 37(1):145--151.

\bibitem[{Liu et~al.(2019{\natexlab{a}})Liu, Jiang, He, Chen, Liu, Gao, and Han}]{liu2019variance}
Liyuan Liu, Haoming Jiang, Pengcheng He, Weizhu Chen, Xiaodong Liu, Jianfeng Gao, and Jiawei Han. 2019{\natexlab{a}}.
\newblock On the variance of the adaptive learning rate and beyond.
\newblock \emph{arXiv preprint arXiv:1908.03265}.

\bibitem[{Liu et~al.(2019{\natexlab{b}})Liu, Ott, Goyal, Du, Joshi, Chen, Levy, Lewis, Zettlemoyer, and Stoyanov}]{liu2019roberta}
Yinhan Liu, Myle Ott, Naman Goyal, Jingfei Du, Mandar Joshi, Danqi Chen, Omer Levy, Mike Lewis, Luke Zettlemoyer, and Veselin Stoyanov. 2019{\natexlab{b}}.
\newblock Roberta: A robustly optimized bert pretraining approach.
\newblock \emph{arXiv preprint arXiv:1907.11692}.

\bibitem[{Mihaylov et~al.(2018)Mihaylov, Clark, Khot, and Sabharwal}]{mihaylov2018can}
Todor Mihaylov, Peter Clark, Tushar Khot, and Ashish Sabharwal. 2018.
\newblock Can a suit of armor conduct electricity? a new dataset for open book question answering.
\newblock In \emph{Proceedings of the 2018 Conference on Empirical Methods in Natural Language Processing}, pages 2381--2391.

\bibitem[{Mihaylov and Frank(2018)}]{mihaylov2018knowledgeable}
Todor Mihaylov and Anette Frank. 2018.
\newblock Knowledgeable reader: Enhancing cloze-style reading comprehension with external commonsense knowledge.
\newblock In \emph{Proceedings of the 56th Annual Meeting of the Association for Computational Linguistics (Volume 1: Long Papers)}, pages 821--832.

\bibitem[{Molchanov et~al.(2017)Molchanov, Ashukha, and Vetrov}]{molchanov2017variational}
Dmitry Molchanov, Arsenii Ashukha, and Dmitry Vetrov. 2017.
\newblock Variational dropout sparsifies deep neural networks.
\newblock In \emph{International conference on machine learning}, pages 2498--2507. PMLR.

\bibitem[{Pan et~al.(2019)Pan, Sun, Yu, Chen, Ji, Cardie, and Yu}]{pan2019improving}
Xiaoman Pan, Kai Sun, Dian Yu, Jianshu Chen, Heng Ji, Claire Cardie, and Dong Yu. 2019.
\newblock Improving question answering with external knowledge.
\newblock In \emph{Proceedings of the 2nd Workshop on Machine Reading for Question Answering}, pages 27--37.

\bibitem[{Pascanu et~al.(2013)Pascanu, Mikolov, and Bengio}]{pascanu2013difficulty}
Razvan Pascanu, Tomas Mikolov, and Yoshua Bengio. 2013.
\newblock On the difficulty of training recurrent neural networks.
\newblock In \emph{International conference on machine learning}, pages 1310--1318. Pmlr.

\bibitem[{Petroni et~al.(2019)Petroni, Rockt{\"a}schel, Riedel, Lewis, Bakhtin, Wu, and Miller}]{petroni2019language}
Fabio Petroni, Tim Rockt{\"a}schel, Sebastian Riedel, Patrick Lewis, Anton Bakhtin, Yuxiang Wu, and Alexander Miller. 2019.
\newblock Language models as knowledge bases?
\newblock In \emph{Proceedings of the 2019 Conference on Empirical Methods in Natural Language Processing and the 9th International Joint Conference on Natural Language Processing (EMNLP-IJCNLP)}, pages 2463--2473.

\bibitem[{Schlichtkrull et~al.(2018)Schlichtkrull, Kipf, Bloem, Van Den~Berg, Titov, and Welling}]{schlichtkrull2018modeling}
Michael Schlichtkrull, Thomas~N Kipf, Peter Bloem, Rianne Van Den~Berg, Ivan Titov, and Max Welling. 2018.
\newblock Modeling relational data with graph convolutional networks.
\newblock In \emph{The Semantic Web: 15th International Conference, ESWC 2018, Heraklion, Crete, Greece, June 3--7, 2018, Proceedings 15}, pages 593--607. Springer.

\bibitem[{Schlichtkrull et~al.(2021)Schlichtkrull, Cao, and Titov}]{schlichtkrull2021interpreting}
Michael~Sejr Schlichtkrull, Nicola~De Cao, and Ivan Titov. 2021.
\newblock \href {https://openreview.net/forum?id=WznmQa42ZAx} {Interpreting graph neural networks for {\{}nlp{\}} with differentiable edge masking}.
\newblock In \emph{International Conference on Learning Representations}.

\bibitem[{Speer et~al.(2017)Speer, Chin, and Havasi}]{speer2017conceptnet}
Robyn Speer, Joshua Chin, and Catherine Havasi. 2017.
\newblock Conceptnet 5.5: An open multilingual graph of general knowledge.
\newblock In \emph{Proceedings of the AAAI conference on artificial intelligence}, volume~31.

\bibitem[{Srivastava et~al.(2014)Srivastava, Hinton, Krizhevsky, Sutskever, and Salakhutdinov}]{srivastava2014dropout}
Nitish Srivastava, Geoffrey Hinton, Alex Krizhevsky, Ilya Sutskever, and Ruslan Salakhutdinov. 2014.
\newblock Dropout: a simple way to prevent neural networks from overfitting.
\newblock \emph{The journal of machine learning research}, 15(1):1929--1958.

\bibitem[{Sun et~al.(2022)Sun, Shi, Qi, and Zhang}]{sun2022jointlk}
Yueqing Sun, Qi~Shi, Le~Qi, and Yu~Zhang. 2022.
\newblock Jointlk: Joint reasoning with language models and knowledge graphs for commonsense question answering.
\newblock In \emph{Proceedings of the 2022 Conference of the North American Chapter of the Association for Computational Linguistics: Human Language Technologies}, pages 5049--5060.

\bibitem[{Talmor et~al.(2019)Talmor, Herzig, Lourie, and Berant}]{talmor2019commonsenseqa}
Alon Talmor, Jonathan Herzig, Nicholas Lourie, and Jonathan Berant. 2019.
\newblock Commonsenseqa: A question answering challenge targeting commonsense knowledge.
\newblock In \emph{Proceedings of the 2019 Conference of the North American Chapter of the Association for Computational Linguistics: Human Language Technologies, Volume 1 (Long and Short Papers)}, pages 4149--4158.

\bibitem[{Veličković et~al.(2018)Veličković, Cucurull, Casanova, Romero, Liò, and Bengio}]{veličković2018graph}
Petar Veličković, Guillem Cucurull, Arantxa Casanova, Adriana Romero, Pietro Liò, and Yoshua Bengio. 2018.
\newblock \href {https://openreview.net/forum?id=rJXMpikCZ} {Graph attention networks}.
\newblock In \emph{International Conference on Learning Representations}.

\bibitem[{Wang et~al.(2021)Wang, Zhang, Yang, Song, and Qin}]{wang2021gnn}
Kuan Wang, Yuyu Zhang, Diyi Yang, Le~Song, and Tao Qin. 2021.
\newblock Gnn is a counter? revisiting gnn for question answering.
\newblock In \emph{International Conference on Learning Representations}.

\bibitem[{Wang et~al.(2022)Wang, Wu, Zhang, Feng, He, and Chua}]{wang2022reinforced}
Xiang Wang, Yingxin Wu, An~Zhang, Fuli Feng, Xiangnan He, and Tat-Seng Chua. 2022.
\newblock Reinforced causal explainer for graph neural networks.
\newblock \emph{IEEE Transactions on Pattern Analysis and Machine Intelligence}, 45(2):2297--2309.

\bibitem[{Wang et~al.(2019)Wang, Kapanipathi, Musa, Yu, Talamadupula, Abdelaziz, Chang, Fokoue, Makni, Mattei et~al.}]{wang2019improving}
Xiaoyan Wang, Pavan Kapanipathi, Ryan Musa, Mo~Yu, Kartik Talamadupula, Ibrahim Abdelaziz, Maria Chang, Achille Fokoue, Bassem Makni, Nicholas Mattei, et~al. 2019.
\newblock Improving natural language inference using external knowledge in the science questions domain.
\newblock In \emph{Proceedings of the AAAI Conference on Artificial Intelligence}, volume~33, pages 7208--7215.

\bibitem[{Wiegreffe and Pinter(2019)}]{wiegreffe-pinter-2019-attention}
Sarah Wiegreffe and Yuval Pinter. 2019.
\newblock \href {https://doi.org/10.18653/v1/D19-1002} {Attention is not not explanation}.
\newblock In \emph{Proceedings of the 2019 Conference on Empirical Methods in Natural Language Processing and the 9th International Joint Conference on Natural Language Processing (EMNLP-IJCNLP)}, pages 11--20, Hong Kong, China. Association for Computational Linguistics.

\bibitem[{Yasunaga et~al.(2021)Yasunaga, Ren, Bosselut, Liang, and Leskovec}]{yasunaga2021qa}
Michihiro Yasunaga, Hongyu Ren, Antoine Bosselut, Percy Liang, and Jure Leskovec. 2021.
\newblock Qa-gnn: Reasoning with language models and knowledge graphs for question answering.
\newblock In \emph{Proceedings of the 2021 Conference of the North American Chapter of the Association for Computational Linguistics: Human Language Technologies}, pages 535--546.

\bibitem[{Yuan et~al.(2022)Yuan, Yu, Gui, and Ji}]{yuan2022explainability}
Hao Yuan, Haiyang Yu, Shurui Gui, and Shuiwang Ji. 2022.
\newblock Explainability in graph neural networks: A taxonomic survey.
\newblock \emph{IEEE transactions on pattern analysis and machine intelligence}, 45(5):5782--5799.

\bibitem[{Zhang et~al.(2022)Zhang, Bosselut, Yasunaga, Ren, Liang, Manning, and Leskovec}]{zhang2022greaselm}
X~Zhang, A~Bosselut, M~Yasunaga, H~Ren, P~Liang, C~Manning, and J~Leskovec. 2022.
\newblock Greaselm: Graph reasoning enhanced language models for question answering.
\newblock In \emph{International Conference on Representation Learning (ICLR)}.

\bibitem[{Zhao et~al.(2023)Zhao, Luo, Zhang, and Wang}]{zhao2023towards}
Tianxiang Zhao, Dongsheng Luo, Xiang Zhang, and Suhang Wang. 2023.
\newblock Towards faithful and consistent explanations for graph neural networks.
\newblock In \emph{Proceedings of the Sixteenth ACM International Conference on Web Search and Data Mining}, pages 634--642.

\end{thebibliography}
\clearpage
\appendix
\section{LKDA Algorithm}
\label{LKDA algo}
\begin{algorithm}
\caption{LKDA Training and Explanation Process}
\begin{algorithmic}[1]
\Require Text $\boldsymbol{s} = [q; a]$, background subgraph $\mathcal{G}$, model $\mathcal{M}$
\Ensure Faithful explanations from graph encoder $E_{\mathrm{KG}}$
\State \textbf{Input:} Question $q$, Answer $a \in C$, Subgraph $\mathcal{G}$
\State \textbf{Initialize:} Language model encoder $\mathrm{LM}$, Graph encoder $E_{\mathrm{KG}}$

\State \textbf{Step 1: Text and Graph Encoding}
\State Use language model encoder to generate text representations $Z_{\mathrm{LM}} \gets \mathrm{LM}(s)$
\State Use graph attention encoder to generate graph embeddings $E_{\mathrm{KG}} \gets f_{\mathrm{G}}(\mathcal{G})$

\State \textbf{Step 2: Fusion and Masking}
\State Combine $Z_{\mathrm{LM}}$ and $E_{\mathrm{KG}}$ in fusion module $\mathbb{F}$ to generate joint representation
\State Mask text representation to create $\mathcal{M}_{\setminus \boldsymbol{z}}$
\State Calculate target prediction distribution $P(Y|\mathcal{M}_{\setminus \boldsymbol{z}})$

\State \textbf{Step 3: Alignment and Optimization}
\State Minimize Jensen-Shannon divergence $J$ between $P(Y|\mathcal{M})$ and $P(Y|\mathcal{M}_{\setminus \boldsymbol{z}})$
\State Use joint objective $\mathbb{L}$ that includes both $J$-based and cross-entropy terms
\State Update model parameters $\theta_t \gets \theta_t - \nabla_{\theta_t} \mathbb{L}$

\State \textbf{Step 4: Post-hoc Explanations}
\State Derive explanations from trained graph encoder $E_{\mathrm{KG}}$
\State Analyze attention weights $\alpha^{h,M}_{ij}$ to identify key semantic relationships in $\mathcal{G}$
\State Align post-hoc explanations with graph encoder's training

\State \textbf{Output:} Faithful explanations indicating reasoning process of model $\mathcal{M}$

\end{algorithmic}
\end{algorithm}

\section{Graph Neural Network Modeled Knowledge Graph Encoding}
\label{gnn}

The graph encoder \( f_\mathrm{G} \) processes the subgraph \( \mathcal{G} \) by assigning initial embeddings \( \{v_1^{(0)}, \ldots, v_J^{(0)}\} \) to the graph's nodes using pre-trained embeddings. In each GNN layer, these embeddings \( \{v_0^{(\ell-1)}, v_1^{(\ell-1)}, \ldots, v_J^{(\ell-1)}\} \) are updated through information exchange among nodes, leading to updated node embeddings for each entity. Here, \( v_0 \) typically represents the context node:
\begin{equation}
\begin{aligned}
\{v_0'^{(\ell)}, \ldots, v_J'^{(\ell)}\} &= f_\mathrm{G}(\{v_0^{(\ell-1)}, \ldots, v_J^{(\ell-1)}\}) \\
\text{for } \ell &= 1, \ldots, M
\end{aligned}
\end{equation}
This process uses a modified graph attention network (GAT), similar to \citet{yasunaga2021qa}. The GNN calculates node representations \( v_j'^{(\ell)} \) for each node \( v_j \) through message passing:
\begin{equation}
v_j'^{(\ell)} = f_n\left(\sum_{v_s \in \mathcal{N}_{v_j} \cup \{v_j\}} \alpha_{sj} m_{sj}\right) + v_j^{(\ell-1)}
\end{equation}
Here, \( \mathcal{N}_{v_j} \) is the neighborhood of node \( v_j \), \( m_{sj} \) is the message from neighbor \( v_s \) to \( v_j \) and \( f_n \) is a two-layer Multilayer Perceptron (MLP). 
Here, \(\alpha_{sj}\) represents the attention weight between source node \(s\) and target node \(j\). 

\section{Datasets \& KG}
\label{data}

We assess our methods by using two multiple-choice question answering datasets: CommonsenseQA \cite{talmor2019commonsenseqa} and OpenBookQA \cite{mihaylov2018can}, serving as benchmarks for commonsense reasoning.
\paragraph{\textbf{CommonsenseQA.}} A dataset of 12,102 questions in a 5-way multiple-choice format which requires commonsense knowledge beyond mere language understanding. For our experiments, we adopted the \textbf{in-house (IH)} data split by \citet{lin2019kagnet} to facilitate comparison with established baseline methods.
\paragraph{\textbf{OpenBookQA.}} A dataset with its 4-way multiple-choice structure, assesses elementary scientific knowledge through its collection of 5,957 questions, accompanied by a compilation of scientific facts. For this dataset, we relied on the official data splits provided by \citet{mihaylov2018can}.
\paragraph{\textbf{ConceptNet}} \cite{speer2017conceptnet}, a broad knowledge graph, for our tasks. A subgraph $\mathcal{G}$ for each QA context is extracted using the method by \citet{feng2020scalable} with hop size k=2.

\section{Implementation \& Training Details}
\label{hyper}
Our model, following \citet{feng2020scalable, yasunaga2021qa}, includes a 4-head, 5-layer graph encoder (dimension $D=200$) with a 0.2 dropout rate \cite{srivastava2014dropout}. Using RAdam \cite{liu2019variance} with batch size 128, we refine parameters. Input node features from concatenated $[\boldsymbol{q}; \boldsymbol{a}]$ pass through RoBERTa-Large, yielding $1024d$ token embeddings. Gradient clipping at 1.0 \cite{pascanu2013difficulty} and learning rates of $1e^{-5}$ (LM) and $1e^{-3}$ (GNN) are set. Training takes about 2 hours for 30 epochs (10 random seeds) on a 40G A100 GPU, with hyperparameters tuned on the development set.

\end{document}